\documentclass[11pt,a4paper]{article}
\usepackage[hyperref]{acl2018}
\usepackage{times}
\usepackage{latexsym}
\usepackage[whole,autotilde,substmingoth]{bxcjkjatype}

\usepackage{url}

\usepackage{graphics}
\usepackage{algorithm}
\usepackage{algorithmic}
\usepackage{comment}

\aclfinalcopy 

\title{Divide and Generate: Neural Generation of Complex Sentences}

\author{\dag Tomoya Ogata \\\And
  \dag Mamoru Komachi \\
  \dag Tokyo Metropolitan University, \ddag TOYOTA MOTOR CORPORATION\\
  {\tt \{ogata-tomoya1@ed., komachi@\}tmu.ac.jp, tomoya\_takatani@mail.toyota.co.jp} \\\And
  \ddag Tomoya Takatani \\
  } 

\date{}

\begin{document}
\maketitle
\begin{abstract}
We propose a task to generate a complex sentence from a simple sentence in order to amplify various kinds of responses in the database. We first divide a complex sentence into a main clause and a subordinate clause to learn a generator model of modifiers, and then use the model to generate a modifier clause to create a complex sentence from a simple sentence. We present an automatic evaluation metric to estimate the quality of the models and show that a pipeline model outperforms an end-to-end model.
\end{abstract}

\section{Introduction}
In recent years, research on chat dialogue systems has attracted much attention. A typical chat dialogue system selects an appropriate response from a database as an output for a user’s utterance input to it \cite{Ji}.  However, it might not always be possible to find an appropriate response to the user’s utterance if the coverage of the database is limited. Therefore, in order for the system to be able to consistently provide appropriate responses, it is necessary to augment the database beforehand. 

In this research, we address this problem space by providing a method to generate a complex sentence from a simple sentence, through assigning a modifier clause to the simple sentence. Instead of using a manually created corpus for complex sentence generation, we extract a pseudo-parallel corpus for modifier clause generation, and use it to learn a generator model at cheap cost. As shown in the first example of Table \ref{Corpus for modifier}, the input of this method is a sentence that has a modifiable noun phrase, and the output is a sentence with a modifier clause assigned to the input sentence. By doing so, it is possible to augment the database to include a variety of complex sentences based on simple sentences.

The main contribution of this research is as follows.
\begin{itemize}
\item We propose a technique to generate a response by inserting a modifier clause to the input response.
\item We propose a method to automatically create a corpus for inserting a modifier clause from a response database, and to learn a complex sentence generation model with neural networks.
\item We examine evaluation metrics for complex sentence generation, and show that a pipeline method improves both fluency and diversity in comparison with the baseline method, in which the entire generation is done in an end-to-end fashion.
\end{itemize}

\section{Divide and Generate: Neural Generation of a Modifier Clause }
Our objective in this research is to generate various kinds of new responses, by inserting a modifier clause to a simple sentence in a response database to create a complex sentence.
As our generator model for the complex sentences, we use the Encoder-Decoder with Attention \cite{Bahdanau}. In order to train the model, we propose a technique to first automatically build an annotated parallel corpus of pseudo-simple sentences, from a raw corpus of complex sentences. We then present two approaches of learning the generator model: (1) an end-to-end model, which jointly inserts and generates a modifier clause, and (2) a pipeline model, which divides the insertion and generation processes, to guide the generation of a natural modifier clause. 

In Section \ref{Corpus}, we explain how to create a parallel corpus for modifier clauses, and propose a model that inserts a modifier clause in Section \ref{Pipeline model}. The evaluation metric of generated sentences is explained in Section \ref{Evaluation metric}.

\subsection{Pseudo-Simple Corpus} \label{Corpus}
For learning a generation model, it is necessary to have a parallel corpus of complex sentences, in each of which the modifier clause is annotated. However, it is expensive to annotate the corpus manually. Therefore, we create a pseudo-simple corpus by removing the modifier clauses from a raw corpus consisting of complex sentences. We collected training data including modifier clauses and used it for training of complex sentence generation. In this research, we extracted a modifier clause from each complex sentence using Algorithm \ref{extract}.

\begin{algorithm}[t]
\caption{Extraction of a modifier clause \label{extract}}
\begin{algorithmic}
\STATE $chunks \gets $parse the dependency of the sentence
\STATE $\mathit{modifier} \gets \mathit{None}$
\STATE Modifier clause: $\mathit{modifiers} \gets []$
\STATE Modifier clause elements: $\mathit{elements} \gets []$
\FOR{$i = |chunks| - 1$ to 0}
\STATE{$dest \gets chunks[i].dest$}
\IF{$chunks[dest]$ contains ``General Noun" or ``Proper Noun"}
\IF{$chunks[i]$ is a verb phrase or a noun predicate (copula)}
\IF{$\mathit{elements} \neq []$}
\STATE{append $\mathit{elements}$ to $\mathit{modifiers}$}
\ENDIF
\STATE{append $\mathit{chunks[i]}$ to $\mathit{elements}$}
\ENDIF
\ELSIF{$\mathit{dest}$ in $\mathit{elements}$}
\STATE{append $\mathit{chunks[i]}$ to $\mathit{elements}$}
\ENDIF
\ENDFOR
\IF{$|\mathit{modifiers}| >= 1$}
\STATE{$\mathit{modifier} \gets$ random$(\mathit{modifiers})$}
\ENDIF
\STATE return $\mathit{modifier}$
\end{algorithmic}
\end{algorithm}

Note that we can use different corpora for training the generator model and decoding an actual sentence at test time. If the domain of the test corpus is far from that of the training data, unnatural modifier clauses may be generated at test time. Thus, the domain of the training corpus should be carefully chosen.

\subsection{Generator Models} \label {Pipeline model}
The baseline for comparison used in this research is an end-to-end model in which a complex sentence, with a modifier clause given to an input sentence, is generated in an end-to-end fashion by using Encoder-Decoder. Since the end-to-end model simultaneously detects the position to insert a modifier clause and generates the modifier clause for the input sentence, the task gets complicated and is likely to suffer from data sparseness. 

To overcome the limitations of an end-to-end approach, we propose a pipeline model that generates a modifier clause more robustly, by detecting the insertion position and generating the modifier clause separately.

In the pipeline model, the insertion position is detected by a set of rules, and marked with a special token on the input side. The modifier clause is generated by an Encoder-Decoder trained on the pseudo-parallel corpus that includes special tokens that mark the insertion position. Algorithm \ref{detection} (Appendix) shows the rule-based algorithm for detecting the insertion position. After the insertion position is detected, the Encoder-Decoder model creates a complex sentence by generating a modifier clause hinted by the rule. By using a special token to mark the insertion position, our pipeline model can find the noun to modify and generate a modifier clause easily and robustly. 

As shown in the second example of Table \ref{Corpus for modifier}, the pseudo-parallel corpus of the modifier clause includes a special symbol inserted before and after the word to be modified. 

\begin{table*}[t]
\centering
\scalebox{0.85}{
\begin{tabular}{|c|c|c|} \hline
&Input & Output \\ \hline
Original corpus& \shortstack{{\bf{車}}に乗りました\\ I got on a {\bf{car}}} & \shortstack{\underline{彼に借りた}{\bf{車}}に乗りました\\ I got on a {\bf{car}} \underline{I borrowed from him}} \\ \hline
Marked corpus & \shortstack{\texttt{<ins>} {\bf{方法}} \texttt{</ins>} を探しています \\ I am looking for \texttt{<ins>} {\bf{ways}} \texttt{</ins>}} & \shortstack{\underline{この先に進む}{\bf{方法}}を探しています \\ I am looking for {\bf{ways}} \underline{to move forward}} \\ \hline             
\end{tabular}
}
\caption{Examples of input and output. \label{Corpus for modifier}}
\end{table*}

\subsection{Evaluation Metric} \label{Evaluation metric}
In this research, we generate a complex sentence by inserting a modifier clause to a simple sentence. Since there is no specific correct answer for such a task, it is not appropriate to use an evaluation metric that utilizes a reference sentence, such as BLEU and ROUGE, to evaluate the sentence that is generated. 

The purpose of our complex sentence generation task is to improve diversity in the responses, without compromising on fluency to the extent possible. Although our generator model takes simple sentences as input, our goal is to augment the response database, so the quality of the generated data can be evaluated by looking only at the generated data.

Since it is desirable to have fluent sentences, we use perplexity with the N-gram language model created by the test data to assess the fluency of the generator model. 

In addition, we consider, naturally, that the produced sentence has more information than the original sentence. 
Therefore, we use the number of word types in the sentence as a measure for the amount of information. 

\section{Experiment}
\subsection{Setting}
\paragraph{Corpus.}
We extracted conversation sentences from novels posted on an online forum for sharing Japanese novels \footnote{\url{https://syosetu.com}}. We crawled the site and obtained 2,782,577 sentences as of May, 2017. We then created a pseudo-simple corpus as described in Section \ref{Corpus}. We used CaboCha \cite{Kudo1} for dependency analysis, and MeCab \cite{Kudo2} + IPAdic \footnote{\url{http://chasen.naist.jp/stable/ipadic/}} for morphological analysis. Since a simple sentence is assumed as an input at test time, test and development data consist of simple sentences only.  In order to prevent the generated modifier clause from being biased toward a generic but meaningless clause (e.g. ``that I know''), we kept only one instance with the same modifier clauses in the training data. The training data consists of 95,234 sentences, and the test and development data contain 1,000 sentences each.

We also tested whether the model learned by our corpus can correctly give a modifier clause to sentences in an out-of-domain setting. For the out-domain data,  we used simple sentences taken from a chat dialogue corpus \cite{higashinaka}. This corpus is a typed online dialogue corpus consisting of utterances of a system and a user. In this research, we extracted user utterances and generated modifier clauses for each utterance.

\paragraph{Model.}
We conducted an experiment with an end-to-end model and a pipeline model, respectively. In addition, we examined the kind of output that was generated when beam search was performed with a search width 10 in the pipeline model.
The hyper parameter of the neural networks was experimented with a vocabulary size of 10,000, an embed layer of 512, a hidden layer of 512, and a batch size of 128. The initial value of the word vector was word2vec learned from the training data. The optimization algorithm used was Adagrad and the learning rate was 0.01. Model selection was performed by running epochs up to 20 and selecting the number of epochs for which BLEU achieved the maximum score for each dev set.

\paragraph{Evaluation metric.}
We evaluated generated sentences using an automatic evaluation for each model by perplexity of N-gram (N = 4) language model with the modified Kneser-Ney smoothing and the average of word types. 
In addition, as a manual evaluation, we subjectively evaluated the fluency of 210 sentences randomly sampled from each system.  We performed a pairwise comparison of the output of the two models. 

\subsection{Results}
\paragraph{Quantitative evaluation.}
As shown in Table \ref{evaluation}, perplexity is lower in the pipeline model than in the end-to-end model. This indicates that the pipeline model produces more fluent results. Moreover, since the number of word types is greater in the pipeline model, it can output a larger variety of sentences than the end-to-end model.

\begin{table}[t]
\centering
\scalebox{1.0}{
\begin{tabular}{|c|c|c|c|} \hline
& Perplexity & word types \\ \hline
End-to-end model & 54.9 & 13.85 \\ \hline
Pipeline model & \bf{46.9} & \bf{14.44} \\ \hline
\end{tabular}
}
\caption{Perplexity and the average of word types for each generator model. \label{evaluation}}
\end{table}

As for the subjective evaluation, the end-to-end model won 32 times; the pipeline model won 68 times; and there were 110 ties. These results demonstrate that the pipeline model was able to generate sentences with higher fluency than the end-to-end model.
\paragraph{Output example. }
The output of each model is shown in Table \ref{output} (Appendix). 
As in the first example, the end-to-end model sometimes output a same word redundantly in the modifier clause, whereas the pipeline model did not.
Also, in the second example,  \mbox{``大手" (`` major '')} in the input sentence changes to \mbox{``大型" (`` big '')}, which is similar but unnatural in the end-to-end model.
In addition, as in the third and fifth examples, there were many cases where garbled words appeared in the output of the end-to-end model, resulting in the output sentence becoming syntactically and semantically invalid.
Although the pipeline model did not output such a sentence, as in the fourth and fifth examples, it sometimes inserted a modifier clause at an unnatural position and generated a modifier clause not appropriate for the modified noun.

The output of the pipeline model in an out-of-domain setting is shown in
Table \ref{output_external} (Appendix).
As in the first example, the pipeline model could successfully generate a modifier clause for a word contained in the vocabulary of the model in an out-of-domain context.
On the other hand, when a modified word is out-of-vocabulary, there were many cases in which inappropriate modifier clauses were generated as shown in the third example.
\section{Discussion}
\paragraph{Generation model. }
In both the end-to-end model and the pipeline model, it commonly occured that a generic but meaningless modifier clause was inserted to the input sentence. This problem is related to the evaluation metric for selecting models. In this research, we chose the model where BLEU was the highest, but as for the modifier generation task, it is not always possible for the model with the highest BLEU to produce a fluent and diverse modifier clause. As the learning process proceeds, the generated modifier clause tends to vary, whereas loss and perplexity in the development data continue to rise, so that it is necessary to balance the trade-off between fluency and diversity.

The end-to-end model tends to output words with similar meaning, or garbled words, resulting in the generation of unnatural sentences. This indicates that the information to predict the position of the modifier clause is better kept as a special token, rather than as distributed representation in a hidden layer. It is known that style information is better encoded as a special token \cite{Sennrich, Yamagishi}, and our finding is consistent with previous work.
 
Table \ref{output_beam} (Appendix) shows the results of the top three sentences in beam search with a beam width of 10 in the pipeline model. Each sentence has high fluency and is different in meaning. 
Therefore, it can be possible to avoid outputting sentences which have same words redundantly by imposing a penalty for duplication and re-ranking candidates in a beam.

\paragraph{Evaluation metric. }
We evaluated the diversity of the model by the average number of word types in the generated corpus. In this evaluation metric, although a sentence containing more kinds of words receives a better evaluation score, a longer sentence tends to be over-estimated because there is no penalty on the sentence length. Thus, it is necessary to take the sentence length into consideration, possibly weighted by the number of content words in the sentence. 

\section{Related Work}
There is a thread of research on generating sentences based on training data without any input \cite{Bowman}. Their research is similar to ours in that they generate a sentence according to the probability distribution learned from the data beforehand, but we generate a sentence based additionally on a given input. In other words, since the output can be controlled by the input, we can output a sentence in a specific domain or include a specific keyword.

There are also other studies that delve into the topic of complex sentences \cite{Derr, Sato}. The former discusses when to output complex sentences, and therefore it has a purpose that is different from our research. The latter generates Japanese sentences automatically by providing the frames and specifications of a sentence. In his work, it is necessary to select the frame and provide information regarding a subordinate clause in a sentence, while in our work we automatically predict a subordinate clause suitable for main clauses based on training data and generate a modifier clause. 

\bibliographystyle{acl_natbib}
\bibliography{acl2018}
\appendix
\newpage

\begin{table*}[!hbt]
\scalebox{0.7}{
\begin{tabular}{c|c} 
  End-to-end model& Pipeline model\\ \hline \hline
 \shortstack{流石 ， この 国 の 英雄 と 呼ば れ た 英雄 と いう べき です \\  Indeed, you should be called a hero called a hero of this country } & \shortstack{流石 ， \underline{\bf{この 国 を 救っ て くれ た}} 英雄 と いう べき です な \\ Indeed, It should be called as a hero who saved this country}  \\ \hline
\shortstack{大型 柄 ， 私 が 見 た 分量 ござい ます \\ Large pattern, there is the amount I saw } &  \shortstack{\underline{\bf{この 国 を 守る}} 大手 柄 ， おめでとう ござい ます \\ Congratulations on a great achievment to protect this country} \\ \hline
\shortstack{ケイ ぅ ， 何 か を する コース で ね \\ Kei, in a course to do something} & \shortstack{\underline{\bf{俺 が 持っ て き た}} 手作り チョコ って 名目 で ね \\ In terms of a handmade chocolate brought by me} \\ \hline
\shortstack{ルシエル 様 は この 街 に ある 飛行船 を 見 られ ませ ん でし た か \\ Sir Luciel, haven't you seen the airship in this town ?} & \shortstack{\underline{\bf{私 の よう な}} ルシエル 様 は 飛行船 を 見 られ ませ ん でし た か \\ Sir Luciel, like me, haven't you seen the airship ?} \\ \hline
\shortstack{商業 ランク なんて ， 勝手 に ある 幻 々 する だけ だ よ \\ Commercial rank only does gen-gen that there is arbitrarily}& \shortstack{\underline{\bf{俺 が 連れ て き た}} 商業 施設 なんて ， 苛 々 する だけ だ よ \\ I just get annoyed at the commercial facilities I brought}\\ \hline 
\end{tabular}
}
\caption{Comparison of output of the end-to-end model and the pipeline models (The underlined part of the pipeline model represents the inserted modifier clause). \label{output}} 
\end{table*}

\begin{table*}[!hbt]
\centering
\scalebox{0.7}{
\begin{tabular}{c|c|c} 
  Input& \shortstack{犯人 は ， リフレイア という 娘 だ \\ The criminal is a girl named Refleia} & \shortstack{俺 も できる だけ 早く ， 術 を 完成 さ せる \\ I will also complete the technique as soon as possible} \\ \hline \hline
  beam 1 & \shortstack{それ を 知っ て いる 犯人 は ， リフレイア という 娘 だ \\ The criminal who knows it is a girl named Refleia} & \shortstack{俺 も できる だけ 早く ， 魔力 を 使う 術 を 完成 さ せる \\ I will also complete the technique of using magic as soon as possible}\\ \hline
  beam 2 & \shortstack{私 達 を 襲っ て き た 犯人 は ， リフレイア という 娘 だ \\ The criminal who attacked us is a girl named Refleia} & \shortstack{俺 も できる だけ 早く ， 俺 たち を 倒す 術 を 完成 さ せる \\ I will complete the technique of defeating us as soon as possible}\\ \hline
  beam 3 & \shortstack{私 達 を 倒し た 犯人 は ， リフレイア という 娘 だ \\ The criminal who defeated us is a girl named Refleia}& \shortstack{俺 も できる だけ 早く ， 俺 たち を 守る べき 術 を 完成 さ せる \\ I will also complete the technique to protect us as soon as possible}\\ \hline
\end{tabular}
}
\caption{Three-best output of the pipeline model by beam search. \label{output_beam}}
\end{table*}

\begin{table*}[!hbt]
\centering
\scalebox{0.7}{
\begin{tabular}{c|c}
  Input& Pipeline model\\ \hline \hline
 \shortstack{海 は \texttt{<unk:うきうき>} し ま す ね \\ The sea is exciting}& \shortstack{この 街 に ある 海 は うきうき し ま す ね \\ The sea in this city is exciting} \\ \hline
 \shortstack{\texttt{<unk:野菜>} \texttt{<unk:ジュース>} だけ で す \\ There is only vegetable juice}&\shortstack{俺 が 作っ た 野菜 ジュース だけ で す \\ There is only vegetable juice I made} \\ \hline
\shortstack{\texttt{<unk:ジョギング>}さ れ て る ん で す か \\ Are you jogging ?}& \shortstack{俺 が 作っ た ジョギング さ れ て る ん で す か \\ Are you jogging I made ?} \\ \hline
\end{tabular}
}
\caption{Output of the pipeline model in an out-domain setting (\texttt{<unk:>} represents an unknown word on the input side). \label{output_external} }
\end{table*}

\begin{algorithm*}[!hbt]
\caption{Detection of the insert position \label{detection}}
\begin{algorithmic}
\STATE $chunks \gets  $parse the dependency of the sentence
\STATE Noun index list: $\mathit{noun\_index} \gets []$
\STATE Detection flag: $d \gets \mathit{FALSE}$
\FOR{$i$ = 0 to $|chunks|$ - 1}
\IF{$chunks[i]$ contains a noun}
\STATE{append $i$ to $\mathit{noun\_index}$}
\IF{$chunks[i]$ does not have any verb phrase as one of its children}
\STATE{Mark $chunks[i]$ with a special tag}
\STATE{$d \gets \mathit{TRUE}$}
\STATE break
\ENDIF
\ENDIF
\ENDFOR
\IF{$d \neq \mathit{TRUE}$}
\STATE{$i \gets \min(noun\_index)$}
\STATE{Mark $\mathit{chunks[i]}$ with a special tag}
\ENDIF
\STATE return $\mathit{chunks}$
\end{algorithmic}
\end{algorithm*}

\end{document}